%% file: main.tex
\documentclass[10pt,twocolumn,letterpaper]{article}

\usepackage[pagenumbers]{iccv} 


\definecolor{iccvblue}{rgb}{0.21,0.49,0.74}
\usepackage[pagebackref,breaklinks,colorlinks,allcolors=iccvblue]{hyperref}
\usepackage{float}
\usepackage{multirow}
\usepackage{graphicx}
\usepackage{tabularx}
\usepackage{hyperref}

\def\paperID{2} 
\def\confName{ICCV}
\def\confYear{2025}

\title{Topology-Preserving Image Segmentation \\ with Spatial-Aware Persistent Feature Matching}

\author{Bo Wen\textsuperscript{1}, Haochen Zhang\textsuperscript{1}, Dirk-Uwe G. Bartsch\textsuperscript{2} \\ William R. Freeman\textsuperscript{2}, Truong Q. Nguyen\textsuperscript{1}, Cheolhong An\textsuperscript{1} \\
\\
\textsuperscript{1} Department of Electrical and Computer Engineering, University of California, San Diego \\
\textsuperscript{2} Department of Ophthalmology, University of California, San Diego
}

\begin{document}

\crefname{table}{Table}{Tables}
\crefname{figure}{Fig.}{Figs.}


\maketitle
\thispagestyle{empty}
\input{sec/0_abstract}

\input{sec/1_intro}

\input{sec/2_related_work}
\input{sec/3_method}
\input{sec/4_experiments}

\input{sec/5_conclusions}
{
    \small
    \bibliographystyle{ieeenat_fullname}
    \bibliography{main}
}

\input{sec/X_suppl}

\end{document}

%% file: sec/0_abstract.tex
\begin{abstract}
  Topological correctness is critical for segmentation of tubular structures, which pervade in biomedical images. Existing topological segmentation loss functions are primarily based on the persistent homology of the image. They match the persistent features from the segmentation with the persistent features from the ground truth and minimize the difference between them. However, these methods suffer from an ambiguous matching problem since the matching only relies on the information in the topological space. In this work, we propose an effective and efficient Spatial-Aware Topological Loss Function that further leverages the information in the original spatial domain of the image to assist the matching of persistent features. Extensive experiments on images of various types of tubular structures show that the proposed method has superior performance in improving the topological accuracy of the segmentation compared with state-of-the-art methods. Code is available at \href{https://github.com/JRC-VPLab/SATLoss}{https://github.com/JRC-VPLab/SATLoss}. 
\end{abstract}

%% file: sec/1_intro.tex
\section{Introduction}
\label{sec:intro}

Although great advancement in image segmentation of general objects was made by recent models \citep{SAM,SAM-HQ}, the segmentation of special objects and structures remains a challenge. Tubular structures pervade our world, observed through digital images at various scales-from the twisting cell walls in microscopy view to the meandering rivers and roads in satellite imagery. Such structures usually span the whole image in a net-like pattern and feature many fine details, making them challenging to accurately segment. In general, the segmentation of tubular structures emphasizes the topological correctness over the pixel-level precision. For example, when segmenting the human retinal vessels \citep{Retvesseg_review}, whether the continuity, branching and intersections are preserved is usually more important than whether a peripheral pixel is correctly classified. In major applications, such as automatic estimation of vessel tortuosity \citep{retves_tortuosity} and the artery-vein classification \citep{avcls}, the algorithm performance relies on the vessel topology.

The primary research direction in topology-aware segmentation is the loss function \citep{PTLoss, clDice, WTLoss, DMTLoss, TopoSeg, WTLoss3D, DSCNet, TANet, Persistent_CMR, Topoloss_TPAMI, BMLoss, GraphTopoLoss}. Typically, a topological constraint is used in addition to the pixel classification loss. The topological constraint can be either a direct measure \citep{WTLoss, TopoSeg, WTLoss3D, DSCNet, Persistent_CMR, Topoloss_TPAMI, BMLoss} of the topology of the object or an indirect approximation \citep{PTLoss, clDice, TANet}. The direct methods mostly rely on the persistent homology \citep{CompTopo} of the images and attempt to match and minimize the difference between the persistent features computed from the ground truth and the segmentation. A brief introduction of persistent homology will be provided in \cref{sssec:persistent_homology}. Although many matching algorithms were proposed \citep{WTLoss, TopoSeg, WTLoss3D, DSCNet, Persistent_CMR}, they only rely on the global information in the topological space, i.e., the lifespan of the persistent features, which is ambiguous since the binary ground truth masks yield overlapping persistent features. This could cause more incorrect topological features to be prolonged and correct topological features to be eliminated, making the topological loss function less effective. 

To the best of our knowledge, the only existing work that tried to tackle this issue is the Betti-Matching Loss (BMLoss) \citep{BMLoss}, which enables a more accurate persistent feature matching via an induced matching method \citep{Induced_matching}. However, it requires computing the refined barcodes \citep{BMLoss} and imposes a $\mathcal{O}(n^3)$ time complexity, making this method excessively slow, limiting its applicability to practical tasks. As a matter of fact, the experiments in \citep{BMLoss} can only be run on very small toy datasets with compromised image resolution. 

\begin{figure*}[ht]
  \centering
   \includegraphics[width=0.85\textwidth]{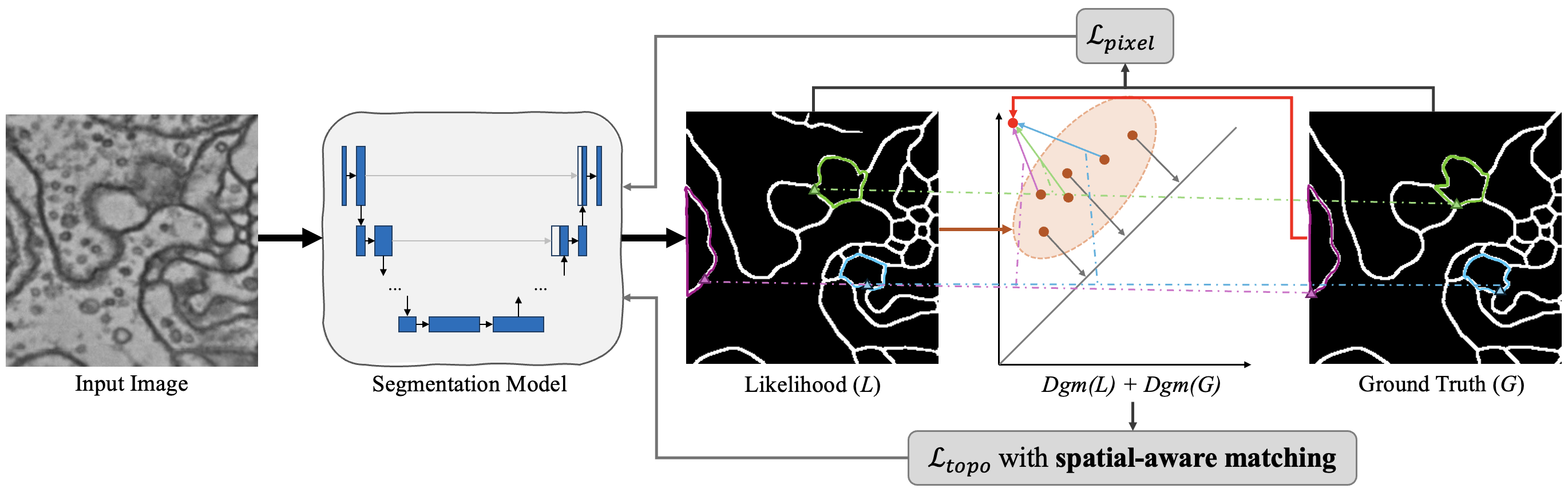}
   \vspace{-0.1cm}
   \caption{Overview of our method. `\textit{Dgm}' is the persistent diagram of the images and the dotted lines indicate the spatial correspondences. Colored triangles indicate creators of topological features and are used as spatial location references for topological feature matching.}
   \label{fig:overview}
   \vspace{-0.4cm}
\end{figure*}

In this paper, we propose a novel Spatial-Aware Topological Segmentation Loss Function (SATLoss) which takes a different approach and directly leverages the spatial relationship between persistent features in the original image in addition to their global topological features (\cref{fig:overview}) for persistent feature matching. While achieving comparable performance with the BMLoss, our method is significantly faster. The introduction of spatial-awareness considerably reduces the incorrect matching between persistent features. As a result, when compared with the other state-of-the-art (SOTA) methods on larger datasets, our proposed method achieves a major improvement in the topological accuracy.

%% file: sec/2_related_work.tex
\section{Related Work}
\label{sec:related_work}

Recent topological segmentation loss functions can be generally divided into indirect methods and direct methods. Indirect methods use certain learned or hand-crafted image features which are proximity of the real topological features while direct methods use strict topological features based on persistent homology. 

For indirect methods, an introductory work \citep{PTLoss} proposed to use an ImageNet pretrained VGG \citep{ImageNet, vgg} network to extract the topological information of the segmentation and ground truth. It is shown that different layers of the extraction network incorporate perceptual features which are approximations of the image topological features. The loss is then computed between the perceptual features of the prediction and the ground truth. \citep{TANet} proposed a topological loss more exclusive for gland segmentation. The loss is computed based on the iterations needed to derive the skeleton of the glands by the erosion operation. \citep{clDice} proposed a clDice loss which approximates the topological features of tubular structures by a relationship between the segmentation and its soft skeleton. It is shown that such a relationship helps strengthen the connectivity within tubular structures, resulting in better topological accuracy in the segmentation.   

\citep{WTLoss} introduces the revolutionary work that for the first time uses the concept of persistent homology in deep image segmentation and opens the door to the direct methods. The idea of all the following works are similar, which is to match the features in the persistent barcodes or persistent diagrams from the prediction and the ground truth and minimize the difference between them. In \citep{WTLoss}, the Wasserstein distance \citep{Wassersteinforpersistence} is used, which later became the most popular matching method. Another work \citep{WTLoss3D} also uses the Wasserstein distance but apply on 3D point cloud data. \citep{TopoSeg} proposed to match the longest persistent barcodes from segmentation with barcodes from ground truth and minimize the difference. The matching result of this method is theoretically the same with the Wasserstein distance on 2D image segmentation task, while this work has a slightly different formulation of the final loss function. \citep{BMLoss} proposed to use an induced matching method \citep{Induced_matching} to solve the ambiguous matching issue due to considering only the global topological features in the previous methods. The general idea of this  method is to construct a common ambient space for the two images which maps the persistent features in between to find a better matching. \citep{DSCNet} proposed to use a Hausdorff distance \citep{Hausdorff_distance} to identify the outlier matchings of persistent diagrams, which focuses on persistent features with anomalous appearing and disappearing time.

Although out of the scope of this paper, improved segmentation networks and post processing methods were also proposed for topology-aware image segmentation. \citep{DSCNet} proposed a Dynamic Snake Convolution, which is a variation of the deformable convolution \citep{DUNet} that learns convolutional kernels into different shapes to better adapt to the features of tubular structures.  \citep{DMTPost} uses the Discrete Morse Theory \citep{CompTopo} to probabilistically estimate the missing tubular structure and refine the original segmentation.

%% file: sec/3_method.tex
\section{Method}
\label{sec:method}

\subsection{Background}
\label{ssec:background}

\begin{figure*}[ht]
  \centering
   \includegraphics[width=\textwidth]{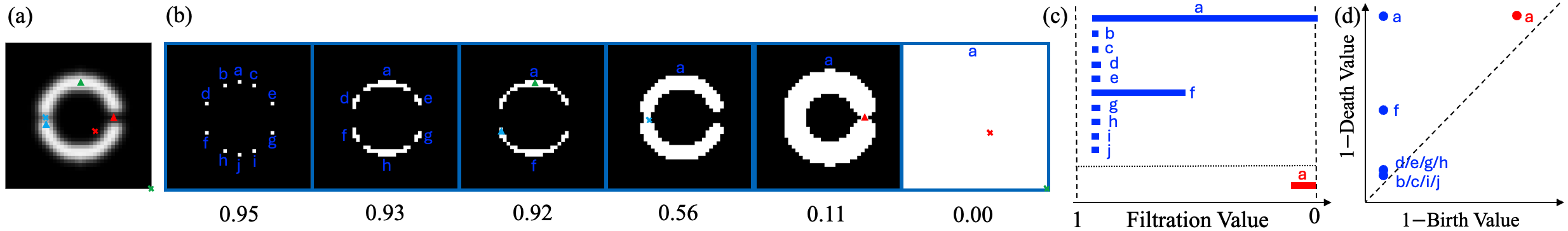}

   \caption{A toy example for persistent homology in digital images: (a) A grayscale image; (b) A visualization of the super-level filtration by representing each sub-complex by its 0-cubes (pixels), the value under each image is the filtration (threshold) value. The triangles and crosses (best zoom in to view) represents the creators and destroyers of selected persistent features (green for 0-a, sky blue for 0-f and red for 1-a), respectively; (c) The persistent barcodes, blue for 0-features and red for 1-feature; (d) The persistent diagram.}
   \vspace{-0.1cm}
   \label{fig:persistent_homology}
   \vspace{-0.3cm}
\end{figure*}

We provide a brief and simplified introduction of related theory and terminologies in computational topology in this part and refer the readers to Edelsbrunner and Harer \citep{CompTopo} for more details and strict definitions. 

\subsubsection{Topology in Digital Images}
\label{sssec:topology}
The topology in digital images are characterized by \textbf{homology groups}. Informally, the p-th homology group is an abelian group that describes the p-th dimensional holes in a topological space. In the context of digital images, the 0-homology group represents the connected components and the 1-homology group represents the loops. Each connected component or loop is called a \textbf{homology feature}. The best way to compute the homology of a digital image is to form the image into a cubical complex. A \textbf{cubical complex} \citep{V-construction, cubical_complex1} is a set of building elements called \textbf{cubes}, each endowed with a dimension. A 0-cubes is a vertex, 1-cube is a line segment between two vertices and 2-cube is a square enclosed by four 1-cubes. In a common \textbf{V-construction} \citep{V-construction} of the cubical complex from a 2D image, the pixels are treated as 0-cubes (vertices). Additionally, an edge that connects two adjacent pixels is treated as a 1-cube and a square enclosed by four edges is treated as a 2-cube. Higher dimensional cubes are usually not considered for 2D images. Additionally, we say that the (n-1)-cubes which encloses the n-cube are the \textbf{faces} of the n-cube, and the n-cube is a \textbf{coface} of the (n-1)-cubes. The n-homology features (a connected component or a loop) can then be determined by the relationships between n-cubes and their cofaces. For example, a connected component can be constructed by a 0-cube and be destructed by filling a 1-cube between it and another 0-cube. Another useful concept is the \textbf{n-Betti number}, which is the rank of the n-homology group. In digital images, 0-Betti number is the number of connected components while 1-Betti number is the number of independent loops.

\begin{figure*}[ht]
  \centering
   \includegraphics[width=0.9\textwidth]{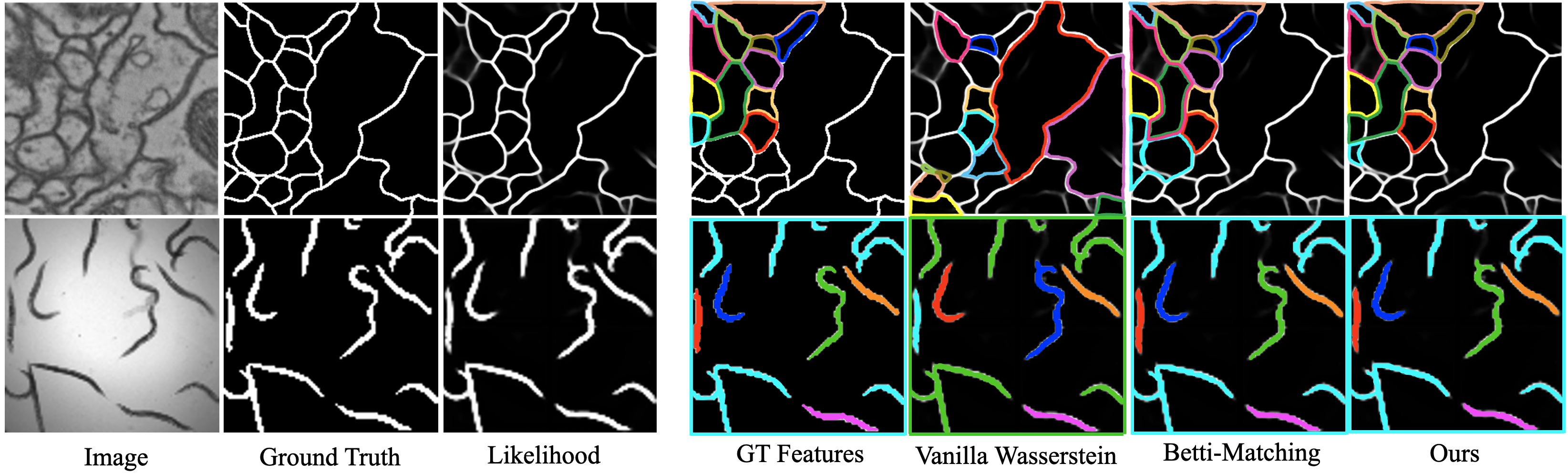}

   \vspace{-0.3cm}
   \caption{Visualization of persistent features matching. A matched 0/1-feature is marked with the same color with the feature in GT.}
   \label{fig:matching}
   \vspace{-0.5cm}
\end{figure*}

\subsubsection{Persistent Homology}
\label{sssec:persistent_homology}

In practice, a digital image $\mathcal{I}$ is filled with varying pixel values. To determine the relationship between cubes then further extract the topology from the image, we need to extend the concept of homology to persistent homology. Consider an image formed into a cubical complex $C$, and a function $f:C \xrightarrow{} \mathbb{R}$ that maps each cube to a value. In practice the values for 0-cubes are simply the pixel values and the values for higher-dimensional cubes are broadcasted from their faces with maximum or minimum values \citep{V-construction}. Furthermore, $f$ is monotonic such that $f(c_{i}) \le f(c_{j})$ if the cube $c_{i}$ is a face of the cube $c_{j}$. Then the sub-level set $C(a) = f^{-1}\left(-\inf, a\right]$ is a sub-complex of the cubical complex $C$. Suppose we have $n$ discrete (pixel) values mapped by $f$, we have $n+1$ different sub-complexes where 
\begin{equation}
\emptyset = C\left(0\right) \subseteq C\left(1\right) \subseteq \dots \subseteq C\left(n\right) = C
\end{equation}
This sequence of complexes is called the \textbf{filtration} of $f$, or to be more specific, the \textbf{sub-level filtration} of $f$. Alternatively, a \textbf{super-level filtration} is the sequence of sub-complexes  considering an $f$ with $f(c_{i}) \ge f(c_{j})$ and the super-level sets $C(a) = f^{-1}\left[a, \inf\right)$. The value $a$ is said to be the \textbf{filtration value}. An easier way to look at it is that, given an image with discrete pixel values in $\left[0, 1\right]$, we threshold the image from 1 to 0. The 0-cubes (pixels) and their cofaces larger than the threshold value are included in the cubical sub-complex. As the threshold value decreases, more cubes are included to form the current sub-complex. Finally, at the smallest threshold value, all cubes from the image are included and the final sub-complex becomes the full cubical complex (\cref{fig:persistent_homology}(b)). A homology feature can be \textbf{born} at a filtration value when a new cube, called the \textbf{creator}, is included into the current sub-complex, and it can be \textbf{killed} at a future filtration value when this feature is destroyed by the inclusion of another higher-dimensional cube, called the \textbf{destroyer}, into a future sub-complex. For example, a pixel whose value is a local maxima gives birth to a 0-homology feature (connected component) at a certain threshold value. As the value decreases, more and more surrounding pixels are included until sometime this 0-feature is connected to another 0-feature in the sub-complex by the introduction of an edge (1-cube) `filled' between two pixels that connects the two areas. Then one of the features is killed (conventionally the newer one). The \textbf{persistent homology} is a method that describes this process where all the homology-features in the image are born and killed in the filtration. A homology-feature that lives in the filtration is said to be a \textbf{persistent feature} and the lifespan of that persistence feature ($born \ filtration \ value - kill \ filtration \ value$) is said to be the \textbf{persistence} of that feature.

\subsubsection{Persistent Barcodes and Diagrams}
\label{sssec:persistent_diagrams}
Persistent barcodes ($\mathcal{B}$) are useful tools to describe and visualize the persistent homology of an image. As shown in \cref{fig:persistent_homology}(c), each barcode represents a persistent feature of the image, where the left edge corresponds to its birth filtration value and the right edge to its death filtration value. The persistent barcodes can be further transformed into a persistent diagram ($\mathcal{D}$, \cref{fig:persistent_homology}(d)). In $\mathcal{D}(\mathcal{I})$, each persistent feature in the image is represented by a dot whose coordinate is $(1 - Birth \ value, 1 - Death \ value)$.

\subsection{Spatial-Aware Topological Loss Function}
\label{ssec:SATLoss}
Recent topology-aware segmentation methods generally follow the idea of matching the persistent diagrams or barcodes between the target and prediction and try to minimize the difference between them. However, (except for BMLoss) the matching is solely based on the distances between persistent features within the diagram or barcodes, i.e., the birth, death values and the persistence of the feature. Given the segmentation model output-likelihood \textit{L}, the binarized prediction \textit{P} and the ground truth \textit{T}, these methods can make \textit{P} to have more similar Betti numbers with the \textit{T}. However, there is no guarantee that each persistent feature from \textit{L} is matched to the correct one from \textit{T}. This issue is prominent in grid-like data. For example, in a 2D image segmentation task, the ground truth mask is usually a binary image, which leads all points in $\mathcal{D}(T)$ to overlap at the upper-left corner of the diagram, i.e., (0, 1). As a result, the distances between each point in $\mathcal{D}(L)$ and any point in $\mathcal{D}(T)$ are the same, which causes confusions when matching the diagrams. We therefore propose to address this problem by introducing spatial awareness that further leverages the spatial distance between two persistent features in the original image when matching persistent diagrams.

\subsubsection{Spatial-Aware Matching and Loss Formulation}
\label{sssec:spatial_awareness}
We introduce the location of creators as a reference for the spatial location of the persistent features. To be more specific, considering a super-level filtration, we compute the normalized (x, y coordinates divided by image height and width, respectively) spatial distance between the creators (if dimension of creator is larger than 0, then use its 0-dimensional face with the largest filtration value) of each pair of persistent features in $\mathcal{D}(L)$ and $\mathcal{D}(T)$. Then, the distance is applied as a spatial weight $s_{p}$ to the distances between two points in the persistent diagram. We follow \citep{WTLoss, WTLoss3D, WT_ref_3} and use the Wasserstein distance \citep{Wassersteinforpersistence}:
\begin{equation}
    W_{q}(\mathcal{D}(L), \mathcal{D}(T)) = [\inf_{\eta :\mathcal{D}(L)\rightarrow \mathcal{D}(T)}\sum_{p\in \mathcal{D}(L)}^{}\left \|p - \eta(p))  \right \|^{q}]^{\frac{1}{q}}
    \label{eq:sa_wa_distance}
\end{equation}
as a base method to compute the best matchings between two persistent diagrams. \(p\) is a persistent feature in $\mathcal{D}(L)$ and \(\eta(p)\) is a candidate matching of \(p\) in $\mathcal{D}(T)$. With the spatial-aware strategy, the modified Wasserstein distance becomes:
\begin{align}
     &W_{q}(\mathcal{D}(L),\mathcal{D}(T))= [\inf_{\eta :\mathcal{D}(L)\rightarrow \mathcal{D}(T)}\sum_{p\in \mathcal{D}(L)}^{} \notag \\
     &\left \|c_{b}(p)-c_{b}(\eta(p))  \right \|^{q}\times \left \|p - \eta(p))  \right \|^{q}]^{\frac{1}{q}}
\label{eq:matching}
\end{align}
where $p \in \mathcal{D}(L)$ and \(c_{b}(p)\), \(c_{b}(\eta(p))\) are the pixel locations of the creator of \(p\) and the creator of its matching in \(T\), respectively. In practice, we take $q=2$ and formulate the SATLoss as:
\begin{equation}
  \mathcal{L}_{topo} = \sum_{p\in\mathcal{D}(L)}^{}s^{*}_{p}([b(p) - b(\eta^{*}(p))]^{2}+[d(p) - d(\eta^{*}(p))]^{2})
\label{eq:loss_define}
\end{equation}
where \(\eta^{*}(p)\) is the optimal matching of \(p\) in \(T\) by \cref{eq:matching}, $b(\cdot)$, $d(\cdot)$ are the birth and death filtration values of the persistent feature and \(s^{*}_{p}=\left \|c_{b}(p)-c_{b}(\eta^{*}(p))  \right \|^{2}\) is the corresponding spatial weight. \cref{eq:loss_define} is a reformulation of \cref{eq:matching} given the matching (spatially-weighted optimal transport) results and the distances in \(\mathcal{D}\) reformulated by the birth and death values of the topological feature. The loss is further performed over the 0 and 1-persistent diagrams and summed. In reality, the number of persistent features in \(L\) is usually larger than that in \(T\) and the redundant persistent features in \(L\) are matched to the diagonal (\cref{fig:overview}; dotted line in \cref{fig:persistent_homology}(d)) of the persistent diagram. In another word:
\begin{equation}
    b(\eta^{*}(p)) = d(\eta^{*}(p)) = \frac{1}{2}(b(p)+d(p))
\end{equation}
Finally, the topological loss function is used with a conventional volumetric (pixel) loss function to jointly optimize the segmentation network:
\begin{equation}
    \mathcal{L} = \mathcal{L}_{pixel} + \lambda\mathcal{L}_{topo}
    \label{eq:total_loss}
\end{equation}
Note that we only consider the location of creators but not destroyers for two reasons: (1) There are persistent features which have infinite persistence thus have no destroyer; (2) For 0-features, the destroyer location of one feature can be close to another 0-feature, which could make the spatial-aware matching less accurate. In comparison, the creator of a 0-feature is the maximum pixel in a feature and the creator of a 1-feature appears at where the loop is closed and can be spatially more informative. Additionally, in implementation we apply a lower bound of 0.05 to \(s_{p}\) to prevent the spatial weight from being over-dominating in the matching.

\cref{fig:matching} shows matching examples between \textit{T} and \textit{L} from a converged segmentation model (more detailed examples in supplementary material (\textit{SM})). From the limited examples that we manually compared, it seems the Betti-matching \citep{BMLoss} is indeed a more accurate method. However, our proposed method still significantly reduces the matching error compared with the Wasserstein matching at a much smaller computational cost than the Betti-matching. Unfortunately, there is no `ground truth' available to compare the matching accuracy numerically. Moreover, it is worth mentioning that although the matching on segmentation from a converged model offers a partial insight on this problem, it does not provide a comprehensive understanding of how the matching precisely influences the whole training process, which can be less straightforward to see. Accordingly, the final segmentation quality on testing images serves as the gold standard for the evaluation and empirically it is shown that a less confused matching enables better topological accuracy in the segmentation (\citep{BMLoss}, \cref{tab:main_tab1}).

\subsubsection{Gradient of SATLoss}
\label{sssec:gradient}
Similar with \citep{WTLoss, BMLoss}, the SATLoss is not differentiable at all pixel locations but only on the \textbf{critical cells} (creators and destroyers) of the persistent features at each iteration. This is because the computation of persistent homology is not differentiable and a surrogate gradient needs to be used. After computing the persistent homology, instead of using the birth and death filtration value from the cubical complex, the pixel values of critical cells are taken from \textit{L} to compute the loss function. This results in the gradient only acting on the critical cell pixels on \textit{L}. By the chain rule, the gradient can be expressed as:
\begin{align}
   \nabla_{\omega}&\mathcal{L}_{topo} = \sum_{p \in \mathcal{D}(L)}^{} 2s^{*}_{p}((L(c_{b}(p))-b(\eta^{*}(p)))\frac{\partial L(c_{b}(p))}{\partial \omega} \notag \\
   & + (L(c_{d}(p))-d(\eta^{*}(p)))\frac{\partial L(c_{d}(p))}{\partial \omega})
\end{align}
where $\omega$ is the model weights, $L(c_{b}(\cdot))$ and $ L(c_{d}(\cdot))$ are the pixel values of the creator and destroyer in \textit{L}. In practice, the gradient pushes the value of creator pixel up and destroyer down if the persistent feature has a match in $\mathcal{D}(T)$ to prolong the persistence of the persistent feature. On contrary, if the persistent feature is matched to the diagonal, the gradient pushes the creator value down and the destroyer value up to shorten the persistence of this feature. The difference with Wasserstein matching is that our SATLoss enables the gradient to act more selectively on the persistent features and their critical cells, reducing the number of persistent features that are incorrectly prolonged or eliminated.

\subsubsection{Computational Cost}
\label{sssec:computation_cost}
Given an $M \times N$ image and $n = M \times N$, our SATLoss has a $\mathcal{O}(n \log (n))$ time complexity, bottle-necked by the step of sorting the cubical complex in computing persistent homology (\textit{PH}). The cost of retrieving spatial location of the critical cells is small, which simply requires mapping the critical cells back to the unsorted cubical complex, and is trivial comparing with the computation of \textit{PH} \citep{Persistence_compute, Persistence_compute2}. In comparison, to address the same \textit{PH} mismatch problem, the BMLoss requires computing the refined barcodes \citep{BMLoss}, where the x-axis values of the barcodes are the indices in the cubical complex instead of filtration values and the computation also requires a reversing of the 1-cubes, which result in a $\mathcal{O}(n^3)$ time complexity. We will show in \cref{ssec:compare_BMLoss} that even when using a very small patch size of $48 \times 48$, there is a huge difference in the runtime. If extended to larger images, the difference can be even much larger (see \textit{SM}).


%% file: sec/4_experiments.tex
\section{Experiments}
\label{sec:experiments}

\begin{table*}[ht]
  \caption{Main results comparing our proposed method with SOTA methods.}
  \vspace{-0.3cm}
  \footnotesize
  \label{tab:main_tab1}
  \centering

  \begin{tabular}{@{}ccc|ccccc@{}}
    \toprule
     \multirow{2}{*}{Dataset} & \multirow{2}{*}{$\mathcal{L}_{pixel}$} & \multirow{2}{*}{$\mathcal{L}_{topo}$} & \multicolumn{3}{c}{Volumetric (\%) \(\uparrow\)} & \multicolumn{2}{c}{Topology \(\downarrow\)} \\
     \cline{4-8}
    & & & \multicolumn{1}{c}{Acc.} & Dice & \multicolumn{1}{c}{clDice} & \(\beta_{0}\) & \(\beta_{1}\) \\

    \midrule

    \multirow{6}{*}{\(CREMI\)} 
        & \textit{bce} & - & \multicolumn{1}{c}{97.75 $\pm$ 0.01} & 84.69 $\pm$ 0.83 & \multicolumn{1}{c}{89.90 $\pm$ 0.22} & 4.547 $\pm$ 0.196 & 1.566 $\pm$ 0.129 \\
        & \textit{bce} & TCLoss & \multicolumn{1}{c}{97.74 $\pm$ 0.04} & 84.96 $\pm$ 0.25 & \multicolumn{1}{c}{89.86 $\pm$ 0.25} & 3.994 $\pm$ 0.278 & 1.475 $\pm$ 0.051 \\
        & \textit{bce} & WTLoss & \multicolumn{1}{c}{97.74 $\pm$ 0.03} & 85.01 $\pm$ 0.19 & \multicolumn{1}{c}{89.98 $\pm$ 0.14} & 3.204 $\pm$ 0.242 & 1.412 $\pm$ 0.083 \\
        & \textit{bce} & He et al. & \multicolumn{1}{c}{97.78 $\pm$ 0.01} & 85.21 $\pm$ 0.14 & \multicolumn{1}{c}{90.13 $\pm$ 0.06} & 3.155 $\pm$ 0.159 & 1.475 $\pm$ 0.030 \\
        & \textit{soft-Dice} & clDice Loss & \multicolumn{1}{c}{97.72 $\pm$ 0.02} & 85.13 $\pm$ 0.17 & \multicolumn{1}{c}{\textbf{90.31 $\pm$ 0.16}} & 2.840 $\pm$ 0.117 & 1.425 $\pm$ 0.011 \\
        & \textit{bce} & \textbf{SATLoss (ours)} & \multicolumn{1}{c}{\textbf{97.78 $\pm$ 0.02}} & \textbf{85.24 $\pm$ 0.13} & \multicolumn{1}{c}{90.16 $\pm$ 0.15} & \textbf{1.534 $\pm$ 0.043} & \textbf{1.386 $\pm$ 0.047} \\

    \midrule

    \multirow{6}{*}{\(DRIVE\)} 
        & \textit{bce} & - & \multicolumn{1}{c}{95.60 $\pm$ 0.48} & 73.54 $\pm$ 3.01 & \multicolumn{1}{c}{73.92 $\pm$ 4.07} & 25.089 $\pm$ 6.216 & 7.667 $\pm$ 1.702 \\
        & \textit{bce} & TCLoss & \multicolumn{1}{c}{95.76 $\pm$ 0.29} & 74.03 $\pm$ 3.92 & \multicolumn{1}{c}{73.91 $\pm$ 4.62} & 23.208 $\pm$ 3.859 & 7.094 $\pm$ 1.773 \\
        & \textit{bce} & WTLoss & \multicolumn{1}{c}{95.94 $\pm$ 0.26} & 75.46 $\pm$ 4.46 & \multicolumn{1}{c}{76.56 $\pm$ 5.70} & 6.352 $\pm$ 1.030 & 6.218 $\pm$ 1.778 \\
        & \textit{bce} & He et al. & \multicolumn{1}{c}{95.85 $\pm$ 0.26} & 74.87 $\pm$ 3.65 & \multicolumn{1}{c}{75.19 $\pm$ 4.69} & 5.199 $\pm$ 1.462 & 6.185 $\pm$ 1.537 \\
        & \textit{soft-Dice} & clDice Loss & \multicolumn{1}{c}{95.06 $\pm$ 0.67} & 74.12 $\pm$ 3.02 & \multicolumn{1}{c}{74.79 $\pm$ 3.78} & 10.193 $\pm$ 2.409 & 6.958 $\pm$ 1.118 \\
        & \textit{bce} & \textbf{SATLoss (ours)} & \multicolumn{1}{c}{\textbf{95.95 $\pm$ 0.23}} & \textbf{75.66 $\pm$ 4.12} & \multicolumn{1}{c}{\textbf{76.69 $\pm$ 5.40}} & \textbf{4.131 $\pm$ 0.315} & \textbf{6.004 $\pm$ 1.546} \\

    \midrule

        \multirow{6}{*}{\(Roads\)} 
      & \textit{bce} & - & \multicolumn{1}{c}{97.61 $\pm$ 0.05} & 58.57 $\pm$ 0.20 & \multicolumn{1}{c}{76.73 $\pm$ 0.27} & 6.015 $\pm$ 0.178 & 0.867 $\pm$ 0.009 \\
      & \textit{bce} & TCLoss & \multicolumn{1}{c}{97.64 $\pm$ 0.05} & 58.21 $\pm$ 0.68 & \multicolumn{1}{c}{76.99 $\pm$ 0.43} & 5.429 $\pm$ 0.326 & 0.865 $\pm$ 0.017 \\
      & \textit{bce} & WTLoss & \multicolumn{1}{c}{97.70 $\pm$ 0.03} & 59.10 $\pm$ 0.32 & \multicolumn{1}{c}{78.59 $\pm$ 0.24} & 2.907 $\pm$ 0.169 & 0.844 $\pm$ 0.013 \\
      & \textit{bce} & He et al. & \multicolumn{1}{c}{97.72 $\pm$ 0.01} & 59.10 $\pm$ 0.14 & \multicolumn{1}{c}{78.93 $\pm$ 0.62} & 2.473 $\pm$ 0.046 & 0.824 $\pm$ 0.039 \\
      & \textit{soft-Dice} & clDice Loss & \multicolumn{1}{c}{97.65 $\pm$ 0.05} & \textbf{60.35 $\pm$ 0.01} & \multicolumn{1}{c}{\textbf{80.94 $\pm$ 0.35}} & 2.887 $\pm$ 0.135 & 0.803 $\pm$ 0.009 \\
      & \textit{bce} & \textbf{SATLoss (ours)} & \multicolumn{1}{c}{\textbf{97.80 $\pm$ 0.02}} & 58.71 $\pm$ 0.50 & \multicolumn{1}{c}{80.90 $\pm$ 0.23} & \textbf{0.562 $\pm$ 0.017} & \textbf{0.720 $\pm$ 0.008} \\

    \midrule

    \multirow{6}{*}{\(CrackTree\)} 
        & \textit{bce} & - & \multicolumn{1}{c}{99.75 $\pm$ 0.16} & 58.37 $\pm$ 9.74 & \multicolumn{1}{c}{64.27 $\pm$ 9.21} & 38.970 $\pm$ 19.890 & 1.357 $\pm$ 0.799 \\
        & \textit{bce} & TCLoss & \multicolumn{1}{c}{99.78 $\pm$ 0.14} & 57.89 $\pm$ 10.57 & \multicolumn{1}{c}{63.00 $\pm$ 10.69} & 33.510 $\pm$ 16.753 & 1.221 $\pm$ 0.886 \\
        & \textit{bce} & WTLoss & \multicolumn{1}{c}{99.75 $\pm$ 0.17} & 57.62 $\pm$ 8.51 & \multicolumn{1}{c}{67.99 $\pm$ 8.66} & 10.944 $\pm$ 7.778 & 1.407 $\pm$ 0.717 \\
        & \textit{bce} & He et al. & \multicolumn{1}{c}{99.76 $\pm$ 0.18} & 58.25 $\pm$ 6.98 & \multicolumn{1}{c}{72.74 $\pm$ 3.93} & 9.044 $\pm$ 7.642 & 1.335 $\pm$ 0.748 \\
        & \textit{soft-Dice} & clDice Loss & \multicolumn{1}{c}{98.03 $\pm$ 3.67} & 38.29 $\pm$ 19.57 & \multicolumn{1}{c}{53.57 $\pm$ 16.43} & $\times$ & $\times$ \\
        & \textit{bce} & \textbf{SATLoss (ours)} & \multicolumn{1}{c}{\textbf{99.77 $\pm$ 0.16}} & \textbf{58.97 $\pm$ 9.42} & \multicolumn{1}{c}{\textbf{73.23 $\pm$ 4.59}} & \textbf{6.191 $\pm$ 3.508} & \textbf{0.901 $\pm$ 0.953} \\

  \bottomrule
  \end{tabular}
  \vspace{-0.2cm}
\end{table*}

\begin{figure*}[ht]
  
  \centering
   \includegraphics[width=0.9\textwidth]{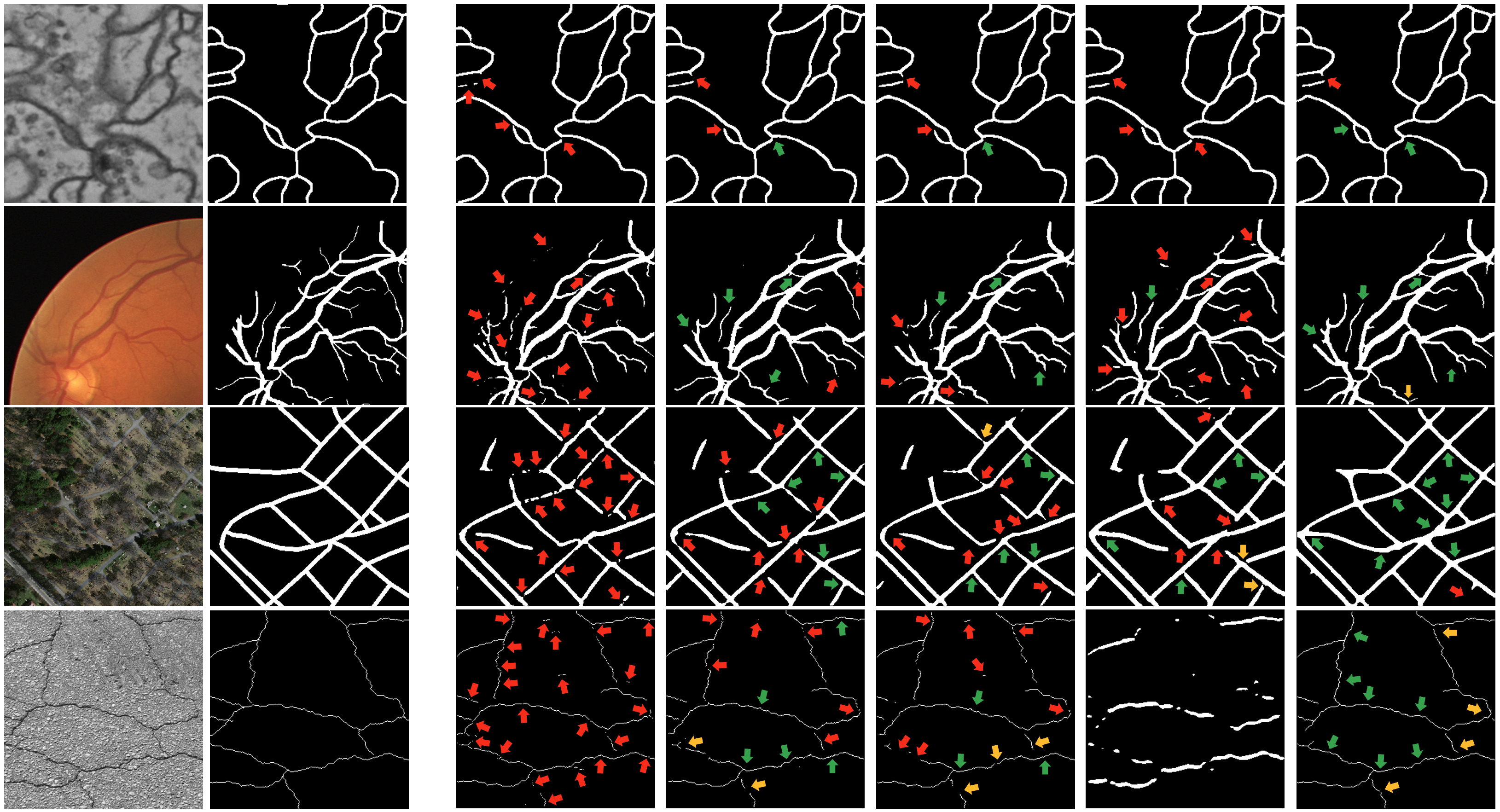}

   \vspace{-0.2cm}
   \caption{Qualitative comparison with SOTA methods. From left to right: image, ground truth, BCELoss, WTLoss, He, clDice, SATLoss. Green, orange and red arrows indicate good, moderate and bad segmentation (topologically), respectively.}
   
   \label{fig:main_qualitative}
   \vspace{-0.4cm}
\end{figure*}

\subsection{Datasets}
\label{ssec:datasets}
Our main comparison with the SOTA topological segmentation loss functions excluding the BMLoss is performed over four diverse datasets and extends to more general tubular structure segmentation beyond biomedical images: 
\begin{itemize}
    \item \textbf{CREMI} \citep{data_cremi}: Segmentation of extracellular matrix of brain neurons in electronic microscopy images. 
    \item \textbf{DRIVE} \citep{data_drive}: Segmentation of human vessels in color fundus retinal images.
    \item \textbf{Roads} \citep{data_road}: Segmentation of various styles of roads in remote images of Massachusetts. 
    \item \textbf{CrackTree} \citep{data_crack} Segmentation of cracks on concrete roads in common photographical images.

\end{itemize}
Considering that the computational cost of persistent homology is still notably higher than conventional segmentation training objectives, it is a common practice in the literature of this field to crop the full image to small image patches for training and computing the evaluation metrics, for example, $64 \times 64$ \cite{clDice, WTLoss}. However, very small patch size leads to limited semantic and topological information. In our experiments, we push it to the limit of our computational resources and use a patch size of around $300 \times 300$ among training, validation and testing. 

\begin{table*}
  \caption{Comparison with BMLoss on C.Elegan-small dataset.}
  \vspace{-0.2cm}
  \scriptsize
  \label{tab:main_tab3}
  \centering

  \begin{tabular}{@{}c|ccccccccc@{}}
    \toprule
     $\mathcal{L}_{topo}$ & \multicolumn{1}{c}{Acc.\(\uparrow\)} & Dice(\textit{P})\(\uparrow\) & \multicolumn{1}{c}{clDice(\textit{P})\(\uparrow\)} & \(\beta_{0}\)(\textit{P})\(\downarrow\) & \(\beta_{1}\)(\textit{P})\(\downarrow\) & Dice(\textit{F})\(\uparrow\) & \multicolumn{1}{c}{clDice(\textit{F})\(\uparrow\)} & \(\beta_{0}\)(\textit{F})\(\downarrow\) & \(\beta_{1}\)(\textit{F})\(\downarrow\) \\

    \midrule

      TCLoss & 98.66$\pm$0.04 & \textbf{90.95$\pm$1.63} & 94.64$\pm$0.84 & 0.470$\pm$0.05 & 0.264$\pm$0.09 & \textbf{92.65$\pm$0.52} & 96.14$\pm$0.50 & 1.517$\pm$0.35 & 0.906$\pm$0.25\\
      
      clDice Loss & 98.59$\pm$0.04 & 90.22$\pm$1.29 & 94.59$\pm$0.83 & 0.481$\pm$0.06 & 0.252$\pm$0.06 & 92.12$\pm$0.53 & 95.92$\pm$0.41 & 1.500$\pm$0.06 & 0.922$\pm$0.10 \\
      
      BMLoss & 98.24$\pm$0.11 & 88.03$\pm$1.28 & 92.16$\pm$0.86 & 0.627$\pm$0.20 & 0.256$\pm$0.10 & 90.17$\pm$0.78 & 94.05$\pm$0.85 & 2.072$\pm$0.92 & 0.917$\pm$0.29 \\
      
      \textbf{SATLoss (ours)} & \textbf{98.67$\pm$0.05} & 90.47$\pm$1.94 & \textbf{94.97$\pm$0.97} & \textbf{0.435$\pm$0.14} & \textbf{0.235$\pm$0.08} & 92.39$\pm$0.60 & \textbf{96.25$\pm$0.31} & \textbf{0.839$\pm$0.19} & \textbf{0.856$\pm$0.23} \\

  \bottomrule
  \end{tabular}
  \vspace{-0.1cm}
\end{table*}

\begin{table*}
  \caption{Comparison with BMLoss on Roads-small dataset.}
  \vspace{-0.2cm}
  \scriptsize
  \label{tab:main_tab2}
  \centering

  \begin{tabular}{@{}c|ccccccccc@{}}
    \toprule
     $\mathcal{L}_{topo}$ & \multicolumn{1}{c}{Acc.\(\uparrow\)} & Dice(\textit{P})\(\uparrow\) & \multicolumn{1}{c}{clDice(\textit{P})\(\uparrow\)} & \(\beta_{0}\)(\textit{P})\(\downarrow\) & \(\beta_{1}\)(\textit{P})\(\downarrow\) & Dice(\textit{F})\(\uparrow\) & \multicolumn{1}{c}{clDice(\textit{F})\(\uparrow\)} & \(\beta_{0}\)(\textit{F})\(\downarrow\) & \(\beta_{1}\)(\textit{F})\(\downarrow\) \\

    \midrule

      TCLoss & 96.10$\pm$0.64 & \textbf{48.66$\pm$5.06} & 63.66$\pm$6.93 & 2.591$\pm$0.87 & 0.487$\pm$0.31 & \textbf{63.73$\pm$5.51} & 68.55$\pm$5.98 & 197.7$\pm$63.0 & 54.7$\pm$27.1\\
      
      clDice Loss & 95.50$\pm$0.52 & 46.71$\pm$3.04 & 66.04$\pm$3.11 & 1.941$\pm$0.94 & 0.488$\pm$0.30 & 61.04$\pm$3.74 & 68.88$\pm$4.15 & 179.4$\pm$85.2 & 52.3$\pm$27.3 \\
      
      BMLoss & 94.91$\pm$0.87 & 46.95$\pm$3.40 & 64.93$\pm$4.83 & 0.896$\pm$0.31 & 0.520$\pm$0.28 & 60.93$\pm$3.69 & \textbf{69.10$\pm$3.65} & \textbf{61.5$\pm$23.1} & \textbf{40.8$\pm$23.0} \\
      
      \textbf{SATLoss (ours)} & \textbf{96.12$\pm$0.53} & 47.73$\pm$3.91 & \textbf{66.17$\pm$3.16} & \textbf{0.815$\pm$0.13} & \textbf{0.481$\pm$0.31} & 63.34$\pm$4.09 & 68.95$\pm$3.95 & 88.9$\pm$17.9 & 51.9$\pm$25.6 \\

  \bottomrule
  \end{tabular}
  \vspace{-0.6cm}
\end{table*}

\begin{table}[h]

  \caption{Comparison of training time with BMLoss, the runtime is measured on an older computer with a Xeno E5-2650 CPU and GTX 1080Ti GPUs since the implementation of BMLoss requires an old PyTorch version which incurs a runtime error during training on newer GPU models. Full comparison is shown in \textit{SM}.}
  \vspace{-0.3cm}
  \footnotesize
  \label{tab:runtime}
  \centering

  \begin{tabular}{@{}c|cc@{}}
    \toprule
     Dataset & SATLoss (ours) & BMLoss \\

    \midrule
     C.Elegan-small & 0.10 $\pm$ 0.01 hours & 6.90 $\pm$ 0.10 hours \\
     Roads-small & 1.60 $\pm$ 0.08 hours & 101.64 $\pm$ 1.82 hours \\
   
  \bottomrule
  \end{tabular}
  \vspace{-0.4cm}
\end{table}

For comparison with the BMLoss, we follow the implementation details in \citep{BMLoss} and use two much smaller datasets with reduced number of samples and downsampled images. Running BMLoss on the previous full datasets is impractical: It is estimated to cost over 20000 hours using an i9-13900K CPU + RTX 4090 GPU to run a one-time training of BMLoss on the Roads dataset, while the same training takes less than 45 hours using our method. The two datasets are picked from the six datasets used in \citep{BMLoss} with the least loss in ground truth quality (especially topological accuracy) caused by the downsampling:
\begin{itemize}
    \item \textbf{C.Elegan-small} \citep{data_elegan}: Segmentation of worm groups in microscopy images. All images in the dataset are used, downsampled from $340 \times 340$ to $96 \times 96$.
    \item \textbf{Roads-small}: About 9\% of the images are picked from the full Roads dataset and downsampled from $1500 \times 1500$ to $375 \times 375$.
\end{itemize}
The images are cropped to a patch size of $48 \times 48$ to further reduce the computational cost. More details on the 6 datasets can be found in \textit{SM}.

\begin{figure}[ht]
  \vspace{-0.3cm}
  \centering
   \includegraphics[width=0.47\textwidth]{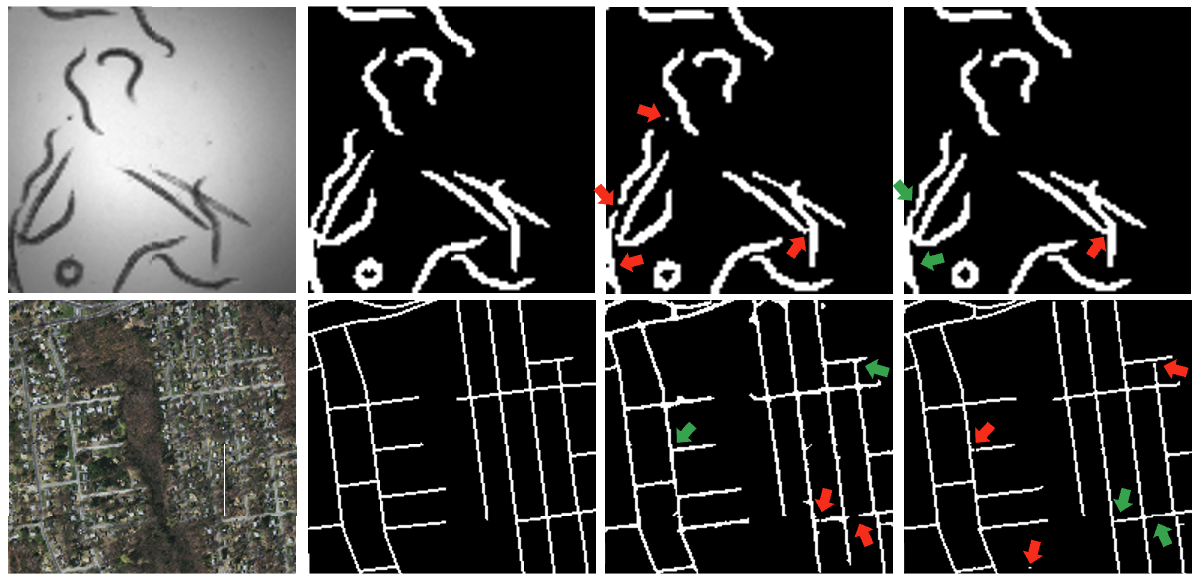}
   \vspace{-0.2cm}
   \caption{Qualitative comparison with BMLoss. From left to right: image; ground truth; BMLoss, SATLoss.}
   \label{fig:BML_qualitative}
   \vspace{-0.6cm}
\end{figure}

\subsection{Implementation Details}
\label{sec:implementation_details}
Since the topic is the loss function, we follow \citep{clDice, WTLoss, BMLoss, TopoSeg, DMTLoss} and adopt a consistent, general U-Net \cite{U-Net} network structure and set of hyperparameters for all experiments. An Adam optimizer is used with a weight decay of 1e-3 and an initial learning rate of 1e-3, decayed to 1e-4 at half of the total number of epochs. We train 30 epochs for Roads, 50 epochs for CREMI and CrackTree and 100 epochs for DRIVE. We use the best pixel loss reported in the corresponding paper for each of the baseline methods and use the BCE loss as pixel loss with our proposed method. The best weights $\lambda$ for the topological loss functions are determined by the ablation studies in \cref{sec:ablations}. All experiments are repeated with 3 random seeds. For the smaller CrackTree and DRIVE datasets, we further adopt a 3-fold cross validation under each seed. In each independent random experiment, the model with the best Dice score on the validation set is used for testing and the average test scores and their standard deviations over these experiments are reported. Our implementation is under the PyTorch framework, where the computation of persistent homology relies on the GUDHI (Geometry Understanding in Higher Dimensions) library \citep{gudhi:urm}. 

For C.Elegan-small and Roads-small datasets, we follow \citep{BMLoss} to determine the network structures and training hyperparameters. We train all loss functions using the same hyperparameters reported in \citep{BMLoss} without finetuning in our method. Additionally, in \citep{BMLoss} experiments were run for only once. Consider the small dataset size, we instead use a 3-fold cross validation where each fold is repeated with 3 random seeds to make the results statistically more convincing.

\subsection{Evaluation Metrics}
\label{sec:metrics}
Two types of metrics: volumetric and topological are used to evaluate the general segmentation quality and the topological accuracy, respectively. For volumetric metrics, we adopt the pixel-wise accuracy and the Dice score. We also use a clDice score \citep{clDice}, which although is a pixel-based metric, indirectly implies the topological correctness. For topological metrics, we use the 0-th and 1-st Betti number error (\(\beta_{0}\), \(\beta_{1}\)), which are the most commonly-used measurements of topological accuracy. We stop using the Betti-matching metric proposed in \citep{BMLoss} because the computational cost is too high given the comprehensive experiments.

\begin{figure*}[ht]
  \centering
   \includegraphics[width=0.9\textwidth]{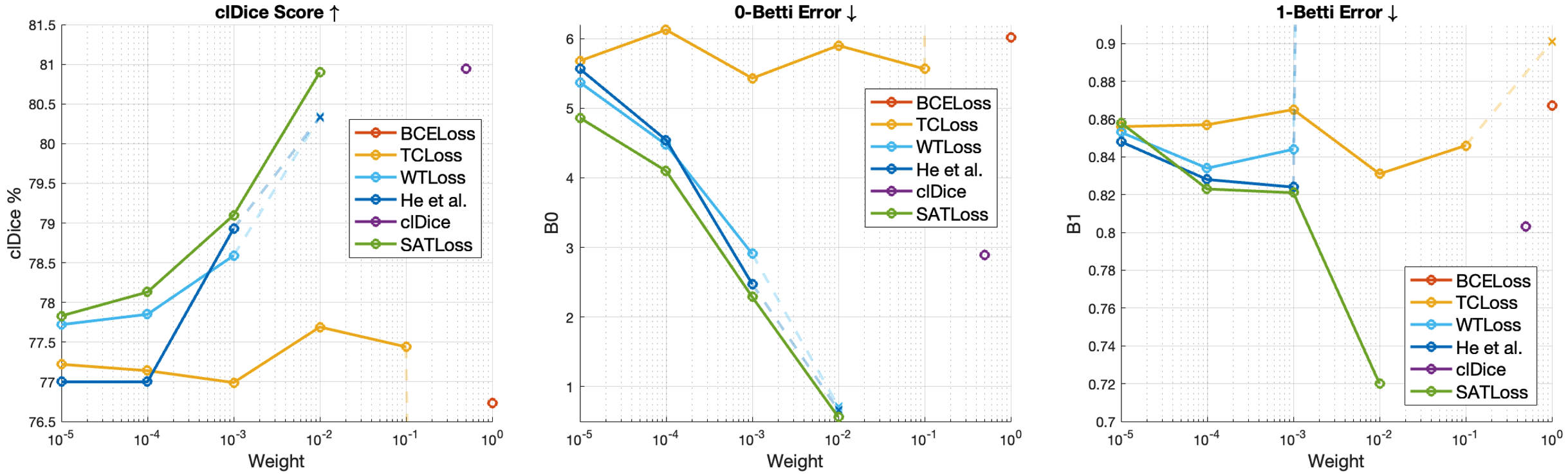}
   \vspace{-0.3cm}
   \caption{Ablation study on weight of topological loss functions on Roads dataset. Results on the other datasets can be found in \textit{SM}.}
   
   \label{fig:main_ablation}
   \vspace{-0.3cm}
\end{figure*}

\subsection{Comparison with SOTA Methods}
\label{sec:compare_SOTA}
For the main experiment performed on the four full datasets, we compare our method with the most commonly used WTLoss \citep{WTLoss} and a most recent work He et al. \citep{TopoSeg} which is a slight variation of the WTLoss. We also compare with the clDice loss \citep{clDice} and another most recent TCLoss \citep{DSCNet} that matches the persistent diagram via a Hausdorff distance. As demonstrated in \cref{tab:main_tab1} and \cref{fig:main_qualitative}, our proposed method shows superior performance in topological accuracy as well as the overall segmentation quality over the SOTA methods. More discussions on the results can be found in \textit{SM}.

\subsection{Comparison with BMLoss}
\label{ssec:compare_BMLoss}
For comparison with BMLoss, we also include the clDice and TCLoss in the baseline, as the first method is the newest baseline in \citep{BMLoss} and TCLoss is a newer method than BMLoss (which did not include BMLoss in their baseline). On another note, although the metrics are commonly computed on the image patches \citep{WTLoss, DSCNet, clDice}, in \citep{BMLoss} they are computed over the full image. We therefore report the results on both the patch size \textit{P} and full image size \textit{F}. The results are shown in \cref{tab:main_tab2}, \cref{tab:main_tab3} and \cref{fig:BML_qualitative}. While our proposed method does not achieve the same performance level in global topological accuracy on full image size with BMLoss on the Roads-small dataset, on boarder evaluations which take the other results into consideration, our method demonstrates comparable, if not superior, performance. Furthermore, as reported in \cref{tab:runtime}, our method achieves the results at a much smaller training time compared with BMLoss. Such a fundamental constraint of BMLoss is much less a problem of our method. It is important to note that in addition to the algorithm complexity, the coding language and implementation details can also influence the training time. The bottom-level codes in the GUDHI library we use to compute persistent homology are written in C++, while the codes for BMLoss are all written in Python. So this comparison is unfair in a strict sense. Given the complexity of the BMLoss, a reimplementation in C++ is challenging and is beyond the scope of this study. To the best of our knowledge, the Python version we used is the only and official implementation available. Therefore, despite the differences in implementation context, our proposed method still demonstrates superior computational efficiency.

\subsection{Ablation Studies on Weight $\lambda$}
\label{sec:ablations}
The results in \cref{tab:main_tab1}, \cref{tab:main_tab3}, \cref{tab:main_tab2} report under the best weight $\lambda$ (\cref{eq:total_loss}) for the topological loss functions studied in this section. We experiment the WTLoss, He et al. and our proposed method using weights from \textit{1e-5} to \textit{1e-2}. For TCLoss whose value is from fewer persistent features, we extend the range to \textit{1e-0}. For clDice and BMLoss, we use the optimal weights since they are reported in their papers (0.3 for clDice on CREMI, 0.5 on other datasets; 0.5 for BMLoss on both datasets). All experiments in the ablation study are also repeated with 3 random seeds and the mean value is reported in the charts. For our method and the two methods without spatial-aware matching (WTLoss and He et al.), as reported in \cref{fig:main_ablation}, we observe that the topological accuracy is higher if a larger $\lambda$ is applied. However, if $\lambda$ is too large, the training will either converges badly or produce results with many small holes on the foreground that significantly increase the 1-Betti error (which we denote by dotted lines and crosses in \cref{fig:main_ablation}). Our proposed method usually allows using a higher weight to further reduce the topological error. This effect is observed in repeated experiments over different datasets. A potential explanation is that the normalized spatial weights decrease the magnitude of gradient of the loss function. Nevertheless, even under the same weight, our method still consistently outperforms the others.

%% file: sec/5_conclusions.tex
\section{Conclusions}
\label{sec:conclusions}

In this paper, we proposed a novel Spatial-Aware Topological Segmentation Loss Function that leverages the spatial information of persistent features in addition to their global topological information to match the persistent diagrams. The introduction of spatial-awareness effectively reduces the incorrect matchings compared with conventional methods and achieves considerably better topological accuracy in the segmentation. When compared with another (and the only) Betti-matching method which tries to tackle the same problem, our proposed method achieves comparable performance at a much lower computational cost. Future works include investigating how to leverage more accurate spatial information of the persistent features without introducing large computational overhead.

%% file: sec/X_suppl.tex
\clearpage
\setcounter{page}{1}
\maketitlesupplementary

\section{Details on Datasets}
\label{sec:D-datasets}
\subsection{CREMI}
The CREMI dataset \citep{data_cremi} (\href{https://cremi.org/data/}{Link}) has three volumes (A, B \& C), volume A, B has 125 scans, volume C has 123 scans. Each scan is an $1250 \times 1250$ 2D image. We use volume A,B and scan 1-14, 16-74, 76-81 in volume C for training, scan 82-92 in volume C for validation and scan 93-125 for testing. Each scan is cropped into 25 $250 \times 250$ patches. In total, there are 8225 for training, 275 for validation and 825 for testing. The original CREMI dataset only provide the ground truth mask for each cell. We take the spaces between the cells to form the ground truth for the cell boundaries (extracellular matrix). The evaluation metrics are computed on the same patch size (the same for the following datasets).

\subsection{DRIVE}
The DRIVE dataset \citep{data_drive} (\href{https://www.kaggle.com/datasets/andrewmvd/drive-digital-retinal-images-for-vessel-extraction}{Link}) has 20 $565 \times 584$ images. We slightly remove the redundant blank margins around the circular effective imaging area and thus crop each image to $550 \times 550$. For fold 1, we use 21-32 for training, 33-34 for validation, 35-40 for testing; For fold 2, we use 29-40 for training, 27-28 for validation and 21-26 for testing; For fold 3, we use 23-28, 35-40 for training, 21-22 for validation and 29-34 for testing. Each image is then cropped into 4 $275 \times 275$ image patches. In total, there are 48 for training, 8 for validation and 24 for testing. 

\subsection{Roads}
The (Massachusetts) Roads dataset \citep{data_road} (\href{https://www.cs.toronto.edu/~vmnih/data}{Link}) has in total 1171 $1500 \times 1500$ images, and is split into 1108 training, 14 validation and 49 testing by the dataset author. We follow this dataset split in our experiments. Each image is cropped into 25 $300 \times 300$ image patches. In total, there are 27700 for training, 350 for validation and 1225 for testing. 

\subsection{CrackTree}
The CrackTree dataset \citep{data_crack} (\href{https://www.kaggle.com/datasets/is4hernandez/cracktree-260}{Link}) has 260 images, most of them have a resolution of $800 \times 600$, while a few others have a resolution of $960 \times 720$. For consistency, we pad all the $800 \times 600$ images to the right to $900 \times 600$ and crop the $960 \times 720$ images from top and bottom (in equal) and right to $900 \times 600$. After sorting all the images by name: For fold 1, we used image 1-200 for training, 201-220 for validation, 221-260 for testing; For fold 2, we used image 1-116, 177-260 for training, 117-136 for validation, 137-176 for testing; For fold 3, we used image 61-260 for training, 1-20 for validation, 21-60 for testing. Each image is then cropped to 6 $300 \times 300$ image patches. In total, there are 1200 for training, 120 for validation and 240 for testing.

\subsection{C.Elegan-small}
We follow as many details as we can in \citep{BMLoss} to prepare the C.Elegan-small and Roads-small dataset. The C.Elegan dataset \citep{data_elegan} (\href{https://bbbc.broadinstitute.org/BBBC010}{Link}) has 200 $696 \times 520$ images. We crop the images and only keep the center $340 \times 340$ effective imaging area and then downsample them to $96 \times 96$. For fold 1, we use A01-C22 for training, C23-D08 for validation and D09-E04 for testing; For fold 2, we use B07-E04 for training, A01-A10 for validation and A11-B06 for testing; For fold 3, we use A01-B16, C23-E04 for training, C13-C22 for validation and B17-C12 for testing. Each image is cropped into 4 $48 \times 48$ patches. In total, there are 280 for training, 40 for validation and 80 for testing.

\subsection{Roads-small}
In total 134 images are selected from the original Roads dataset (mixing training, validation and testing, since we use 3-fold cross validation for this dataset). In \citep{BMLoss}, which images are selected is not provided so we pick image by ourselves. It is difficult to describe the selected images unless providing a long list here but the general rule is to select images with more complexed road networks and less blank regions. The selected images are then divided randomly into training, validation and testing for each fold. The images are then downsampled from $1500 \times 1500$ to $375 \times 375$. Again, details of downsampling are not provided in \citep{BMLoss}, so we use (the same for C.Elegan-small) bicubic interpolation for all images and labels and then binarized the downsampled labels by thresholding them with a value of 0.5. No further manual correction is made to the downsampled labels. Before cropping, the images are padded to the right and bottom to $384 \times 384$. Then, each image is cropped to 64 $48 \times 48$ patches. In total, there are 6400 for training, 640 for validation and 1536 for testing.

\section{Additional Implementation Details}
\label{sec:extra_details}

We train on Roads, CREMI and CrackTree datasets using a batch size of 16 and on DRIVE dataset using batch size of 2 (since bad convergence on larger batch size). During training, we follow \citep{WTLoss} and pad the edges of the image patches with 1 (1 pixel width) to also take the 1-features enclosed by the image boundary into account. We run the main experiments on two computers, one with a i7-12700K CPU + RTX 3090 GPU, another with a i9-13900K CPU + 2 RTX 4090 GPUs. The experiments are randomly distributed into different computers.  For comparison with BMLoss on Roads-small and C.Elegan-small datasets, we run on an older computer with a Xeno E5-2650 CPU + 3 GTX 1080Ti GPUs. This is because the implementation of BMLoss requires an old PyTorch and CUDA version where we find using newer GPU models encounters a random interruption issue during the training. Each experiment only takes one GPU. For implementation of the baseline, we use the official implementation for the respective methods if available (clDice \citep{clDice}: \href{https://github.com/jocpae/clDice}{Link}, BMLoss \citep{BMLoss}: \href{https://github.com/nstucki/Betti-matching/}{Link}). For methods where official implementations are unavailable (TCLoss \citep{DSCNet}, WTLoss \citep{WTLoss}, He et al. \cite{TopoSeg}), we use our own implementation similar with the SATLoss. The computations of persistent homology for these methods are exactly the same, and the difference only exists in the implementation of the loss functions, where we follow the mathematical expressions in the respective papers.

\section{More Discussion on Experimental Results}
\label{sec:result discussion}
One observation in the results that worth discussing is that the TCLoss \citep{DSCNet}, although a most recent method, appears at a very poor position in the baseline. On the one hand, unfortunately, in \citep{DSCNet} the authors did not provide enough implementation details for us to reproduce their results. And they did not provide ablation studies on the topological loss weight, nor did they report the weight used for their method and the baseline methods. On the other hand, the results shown in \citep{DSCNet} seems to have limited improvement over its baselines, including the BCELoss. And in our experiments we observe that the TCLoss has slight improvements over the BCELoss, which are similar with the results in \citep{DSCNet}. In our opinion, the main limitation of the TCLoss is that the use of max and min functions in the Hausdorff distance makes only one pair of persistent features to have gradient from the loss function, which can make the loss less effective. In comparison, the other persistent homology-based methods optimize over all persistent features. We tried to increase the weight for TCLoss to \textit{1e-0} until it diverges to compensate the small loss and gradient value (since they are computed from only one pair of persistent features, although it is an outlier which should usually have larger value than normal pairs). But the effort only seems to provide marginal improvement. Perhaps more investigations are needed to determine the performance and usefulness of this method.

Furthermore, we observe that the clDice loss \citep{clDice} converges badly under some folds and random seeds when applied on the CrackTree dataset. It is potentially because of that the CrackTree dataset has very thin tubular labels (usually only one pixel width). And the labels, according to our examination, do contain inaccuracy and errors. This can make the overlap-based method (both clDice and the soft-Dice it used with) to act poorly. The CrackTree dataset was not used in \citep{clDice}. Our experiments therefore might provide some insights into a limitation of the clDice method.


\section{Additional Results}
\label{sec:extra_results}
\begin{figure*}[h]
  \centering
   \includegraphics[width=\textwidth]{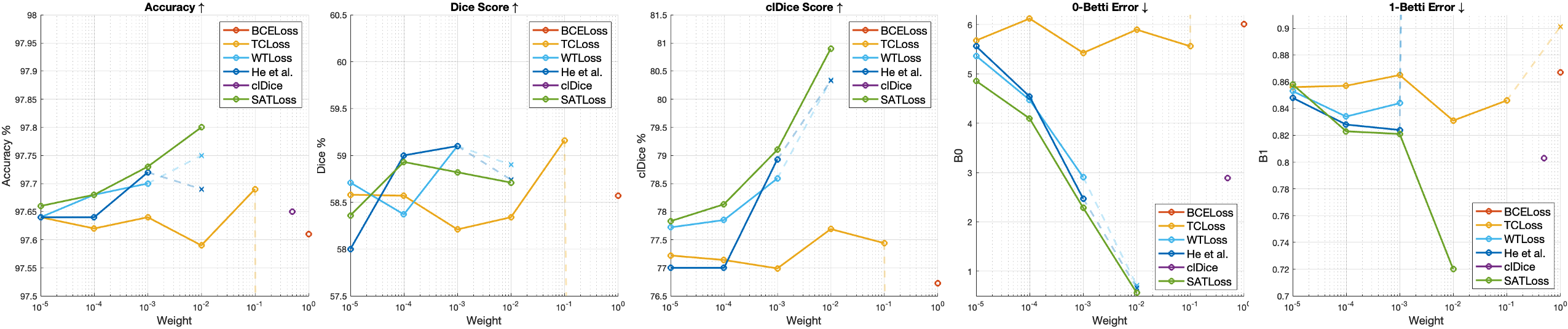}

   \caption{Ablation study results on weight $\lambda$ on Roads dataset (including accuracy and Dice score).}
   \label{fig:ab_roads}
\end{figure*}

\begin{figure*}[h]
  \centering
   \includegraphics[width=\textwidth]{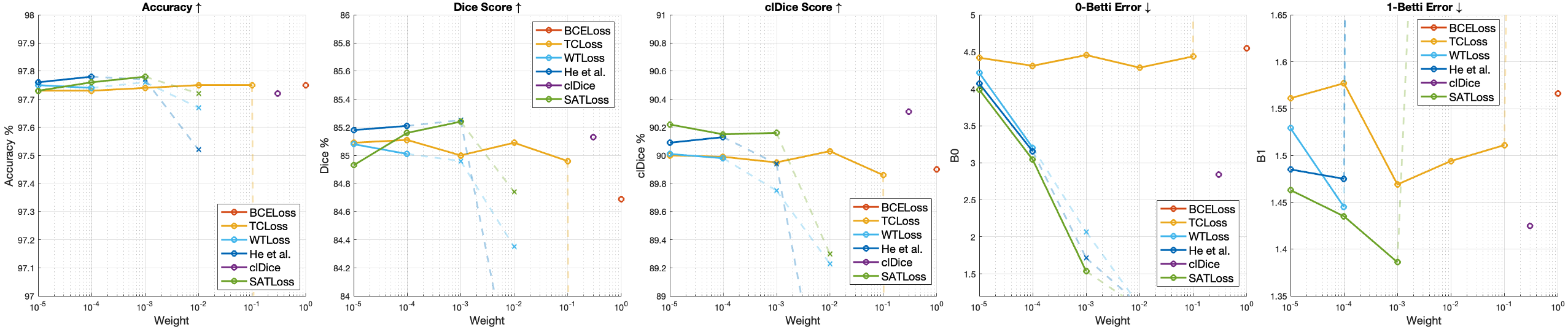}

   \caption{Ablation study results on weight $\lambda$ on CREMI dataset.}
   \label{fig:ab_cremi}
\end{figure*}

\begin{figure*}[h]
  \centering
   \includegraphics[width=\textwidth]{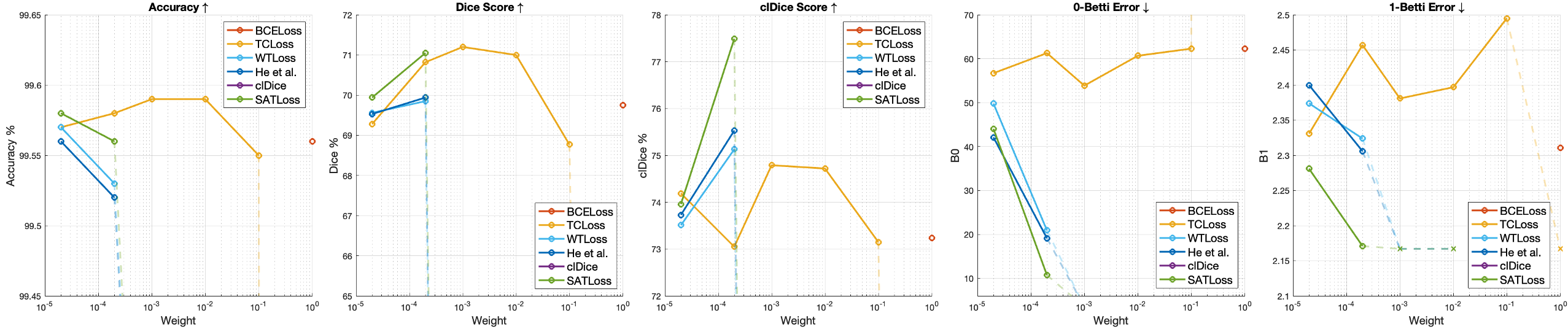}

   \caption{Ablation study results on weight $\lambda$ on CrackTree dataset.}
   \label{fig:ab_cracktree}
\end{figure*}

\begin{figure*}[h]
  \centering
   \includegraphics[width=\textwidth]{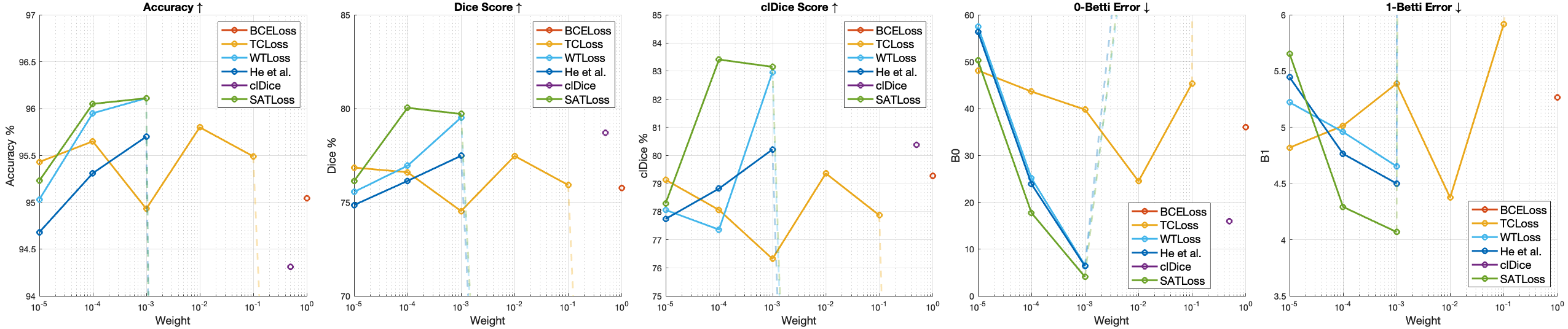}

   \caption{Ablation study results on weight $\lambda$ on DRIVE dataset.}
   \label{fig:ab_drive}
\end{figure*}

\begin{figure*}[h]
  \centering
   \includegraphics[width=\textwidth]{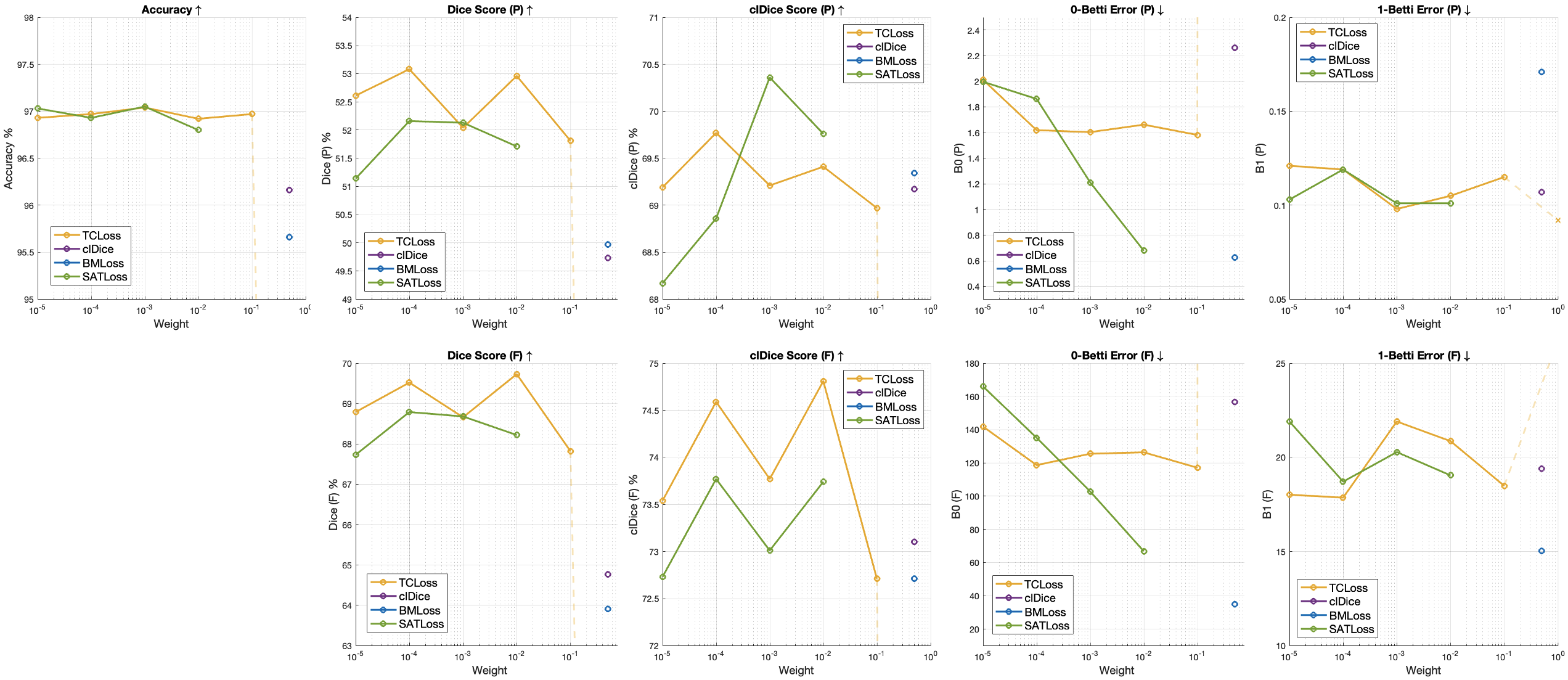}

   \caption{Ablation study results on weight $\lambda$ on Roads-small dataset.}
   \label{fig:ab_roads48}
\end{figure*}

\begin{figure*}[h]
  \centering
   \includegraphics[width=\textwidth]{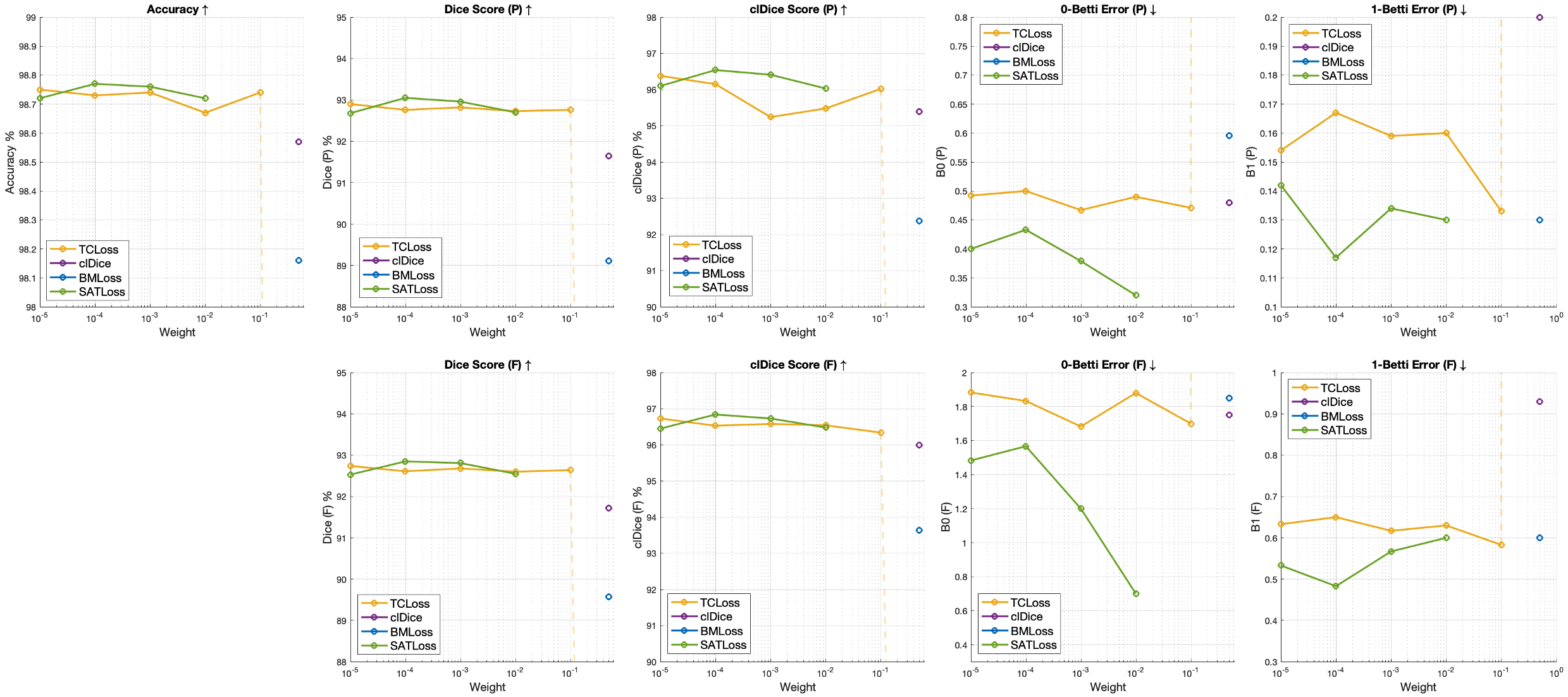}

   \caption{Ablation study results on weight $\lambda$ on C.Elegan-small dataset.}
   \label{fig:ab_elegan}
\end{figure*}

\begin{figure*}[ht]
  \centering
   \includegraphics[width=\textwidth]{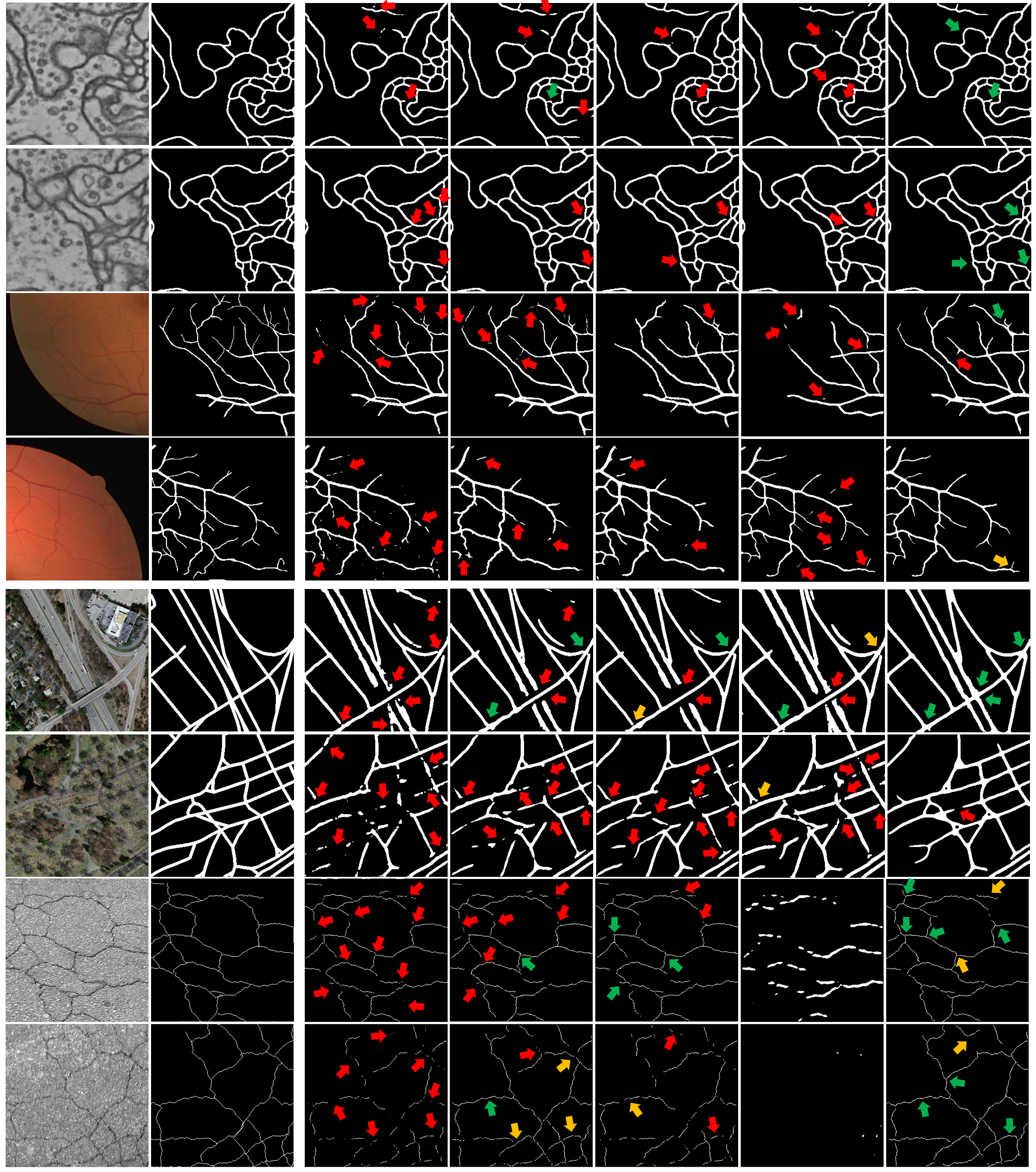}

   \caption{Extra qualitative results on main comparison with SOTA methods. From left to right: image, ground truth, BCELoss, WTLoss, He et al., clDice, SATLoss.}
   \label{fig:extra_qual}
\end{figure*}

\begin{figure*}[ht]
  \centering
   \includegraphics[width=\textwidth]{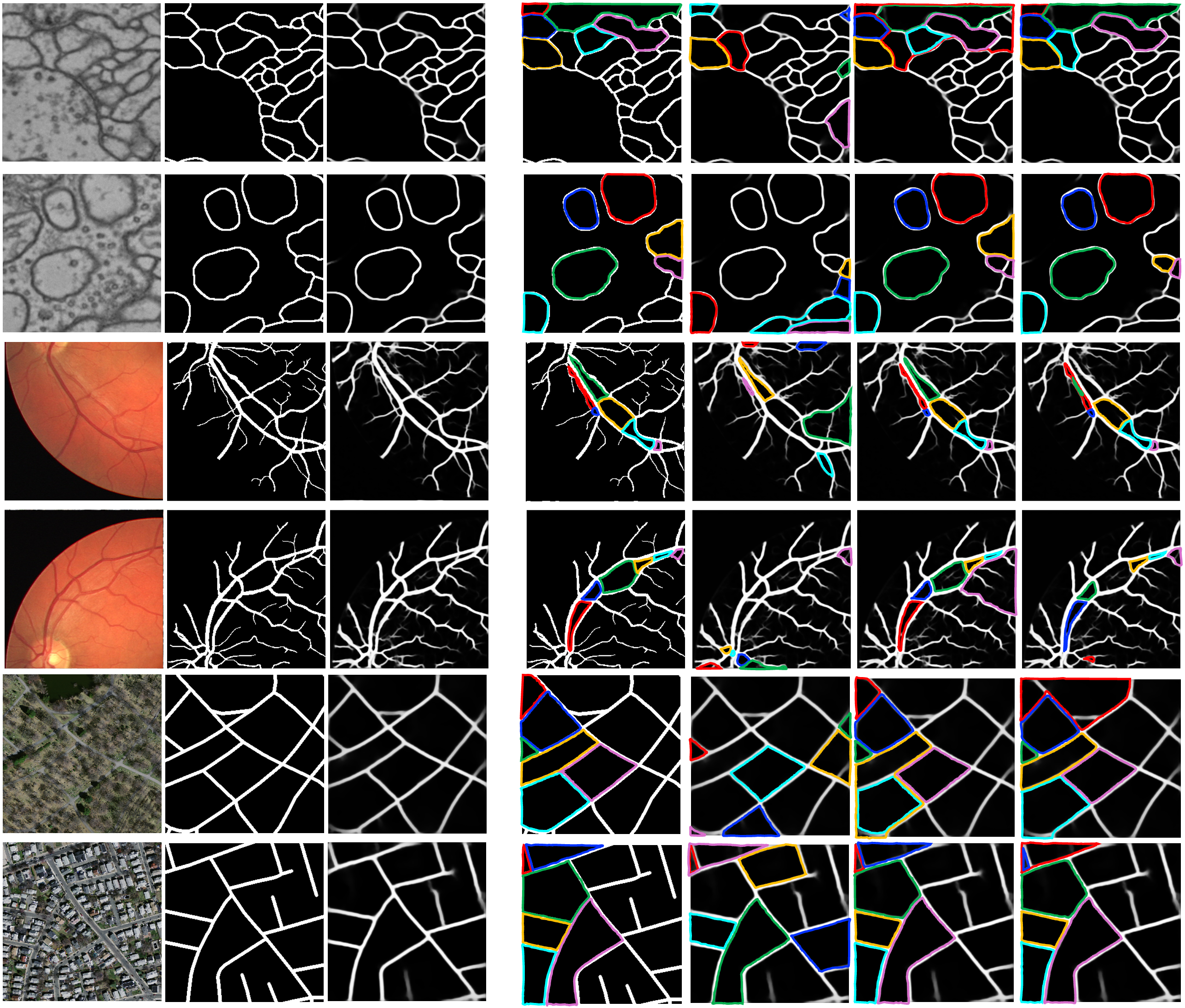}

   \caption{Matching of persistent features from likelihoods predicted by a converged model. Our method usually makes better matching on smaller features, and can make more mistakes on larger and longer features. In general, our method considerably improves the Wasserstein matching (which is usually almost totally messy) at a much lower computational cost than Betti-matching.}
   \label{fig:extra_matching}
\end{figure*}

\begin{figure*}[ht]
  \centering
   \includegraphics[width=\textwidth]{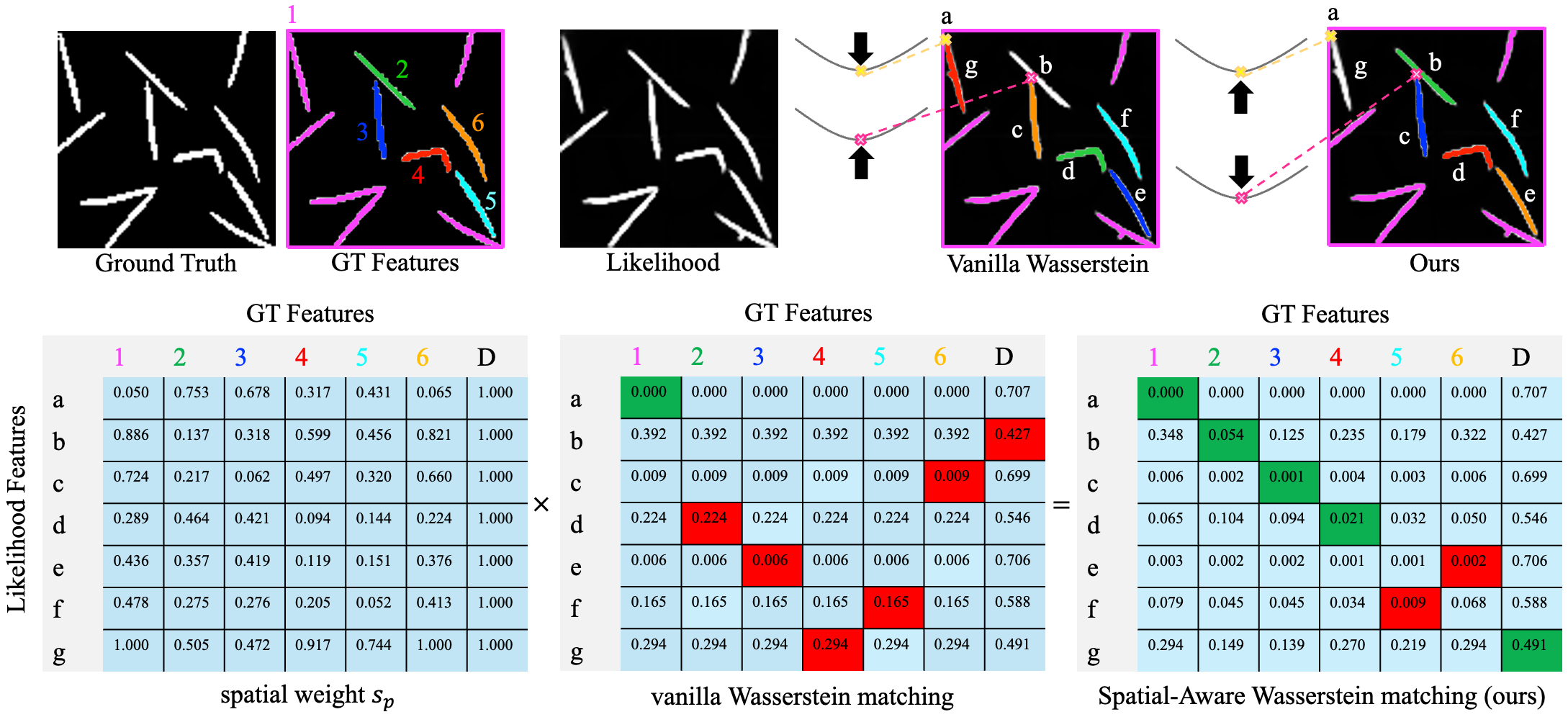}

   \caption{An illustrative example of how the matching works and how spatial-aware matching helps improve topological accuracy in the segmentation. The lower three tables show the spatial weight between each pair of persistent features, the cost matrix used by vanilla Wasserstein matching in previous methods, and the cost matrix after weighted by the spatial weight used by our proposed SATLoss, respectively. The green boxes refer to correct matchings and the red boxes refer to incorrect matchings. D refers to the diagonal of the persistent diagram. Note that by Wasserstein distance (optimal transport) the matching plan giving the least overall cost is selected, to each feature there is no guarantee that the minimal-cost matching is selected. The upper figures shows the matching results and the gradient behavior (see \cref{sssec:gradient} for explanation). At the destroyer of the feature g (yellow cross), by vanilla Wasserstein matching the pixel value is pushed down (since matched with GT features), making feature a and g more separated (which should be connected). Similarly, at the destroyer of the feature b (pink cross), the gradient pushes the value up (since matched with diagonal), making feature b and c connected (which should be separated). In comparison, using the proposed spatial-aware matching, at the destroyer of feature g the pixel value is pushed up, making it more connected to feature a; at the destroyer of feature b the pixel value is pushed down, making it more separated from feature c. Therefore, allowing the segmentation model to make topologically more correct predictions. Similar process occurs repeatedly at different locations on the image in the optimization iterations.}
   \label{fig:illustrative_example}
\end{figure*}

\cref{tab:runtime_full} shows the full training time comparison. The ablation results on the other datasets are provided in \cref{fig:ab_roads}, \cref{fig:ab_cremi}, \cref{fig:ab_cracktree}, \cref{fig:ab_drive}, \cref{fig:ab_roads48}, \cref{fig:ab_elegan}. Due to limited resources and for consistency between datasets, the ablation experiments were only run on the first fold. In addition, we provide more qualitative results in \cref{fig:extra_qual} for our main experiments. In \cref{fig:extra_matching}, we show more persistent feature matching results. In \cref{fig:illustrative_example}, we provide an illustrative example of the motivation of the spatial-aware matching, how it works during the matching process and how it helps with the topological accuracy of the segmentation.

\begin{table}[h]
  \caption{Training time comparison (i9-13900K+RTX 4090)}
  \vspace{-0.2cm}
  \footnotesize
  \label{tab:runtime_full}
  \centering
  \begin{tabular}{@{}c|c|ccccc@{}}
    \toprule
        \textbf{Hours}    & \#iter & clDice & [26] & He. & Ours & BMLoss(Est.)  \\

    \midrule
     Roads & 52k & 2.9 & 44 & 43 & 44 & $>$20000 \\
     CREMI & 26k & 1.0 & 16 & 16 & 16 & $>$8000 \\
     CrackTree & 3.7k & 0.23 & 3.5 & 3.5 & 3.5 & $>$1400 \\
     DRIVE & 2.4k & 0.03 & 0.22 & 0.23 & 0.23 & $>$100 \\
    
  \bottomrule
  \end{tabular}
\end{table}

